\documentclass{acm_proc_article-sp}

\usepackage{graphics,graphicx}
\usepackage{times,amsfonts,amsmath,multirow,url}

\def\R{\mbox{I\hspace{-.15em}R}}

\begin{document}

\title{{\ \\ \LARGE\bf Do not Choose Representation just Change: \\
An Experimental Study in States based EA}}

\author{\begin{tabular}{cccc}
Maroun Bercachi & Philippe Collard & Manuel Clergue & Sebastien Verel \\ \\
\multicolumn{4}{c}{I3S Laboratory - Nice-Sophia Antipolis University - CNRS} \\
\multicolumn{4}{c}{2000 Route Des Lucioles - 06903 Sophia Antipolis - France} \\
\multicolumn{4}{c}{\{bercachi, pc, clergue, verel\}@i3s.unice.fr, http://www.i3s.unice.fr/tea}
\end{tabular}}

\maketitle

\begin{abstract}
Our aim in this paper is to analyse the phenotypic effects (evolvability) of diverse coding conversion operators in an instance of the states based evolutionary algorithm (SEA). Since the representation of solutions or the selection of the best encoding during the optimization process has been proved to be very important for the efficiency of evolutionary algorithms (EAs), we will discuss a strategy of coupling more than one representation and different procedures of conversion from one coding to another during the search. Elsewhere, some EAs try to use multiple representations (SM-GA, SEA, etc.) in intention to benefit from the characteristics of each of them. In spite of those results, this paper shows that the change of the representation is also a crucial approach to take into consideration while attempting to increase the performances of such EAs. As a demonstrative example, we use a two states SEA (2-SEA) which has two identical search spaces but different coding conversion operators. The results show that the way of changing from one coding to another and not only the choice of the best representation nor the representation itself is very advantageous and must be taken into account in order to well-desing and improve EAs execution.
\end{abstract}

\category{G.1.6}{Mathematics of Computing}{Numerical Analysis}{ Optimization} [Stochastic Programming]
\category{I.2.8}{Computing Methodologies}{Artificial Intelligence}{ Problem Solving, Control Methods, and Search} [Heuristic Methods]

\terms{Algorithms, Performance, Experimentation}

\keywords{States based Evolutionary Algorithm, Representation, Coding Coupling, Coding Conversion.}

\section{Introduction}
\begin{figure}
\begin{center}
\begin{tabular}{c}
\includegraphics[width=82mm,height=70mm]{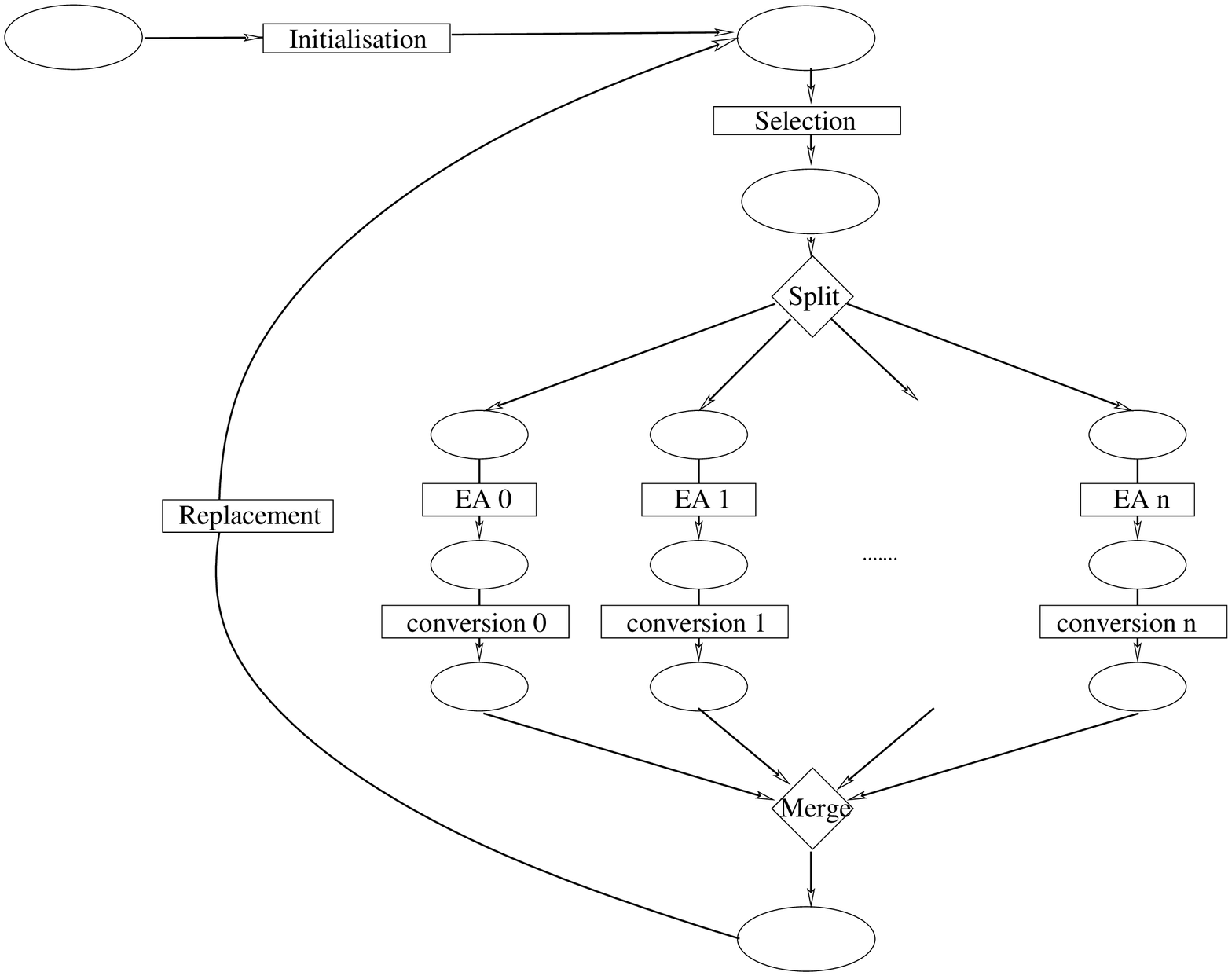}
\end{tabular}
\caption{Scheme of a n-SEA.}
\label{figSEA}
\end{center}
\end{figure}

The choice of the representation of solutions is a very fundamental step and a highly decisive point to take into consideration in EAs functioning. A problem could be difficult for one representation and easy for another one \cite{r21, r42}. It is a challenging task to discover which coding scheme is a suitable one for a specific problem before testing that coding scheme using an evolutionary algorithm (EA). One representation could have a very good behaviour at the begining of the run and a bad one at the end of the run \cite{r19, r18, r42}. Besides, the search bias during genetic search depends on the problem, the structure of the encoded search space and the genetic operators of selection, crossover, and mutation. For every problem there is a large number of possible encodings. It is often possible to follow the principle of minimal alphabets when choosing an encoding for an EA, but simultaneously following the principle of meaningful building blocks can be much harder. This is because our intuition about the structure of the problem space may not translate well in the binary-encoded spaces that EAs expand and spread on \cite{r2, r13, r27, r11, r12}. There are two possible ways of tackling the problem of coding design for meaningful building blocks: 1) Search through possible encodings for a good one while searching for a solution and try to apply the chosen encoding. 2) Incorporate more than one coding scheme simultaneously and change the representation of solutions from one coding to another during the optimization process which can help in well exploring the search space and in increasing the count of building blocks considered meaningful in the solution string. The first choice uses reordering operators, like inversion, that try to look for and then apply the best encoding while searching for the solution. The rest of this paper motivates, develops and illustrates the second approach in use with the states based evolutionary algorithm (SEA). \\
The SEA is a new parallel version of EAs implemented as a group of independent optimization algorithms where each algorithm is considered as a state of a SEA. A state $i$ of a SEA is denoted EA$_i$ and can be any of the optimization algorithms that have been proposed in the literature such as EAs, genetic algorithms (GAs), genetic programming (GP), evolution strategies (ES), etc. An execution of a SEA with $n$ states is equivalent to the execution of $n$ parallel EAs where each EA$_i$ has its own parameter settings (cf. Figure \ref{figSEA}). After each main generation, a SEA contains a $merge$ phase which consists in regrouping all states together in a whole population. During this phase, each state undergoes a mutation to any other state with a given state mutation rate $pMutState$ which can help to maintain diversity over the state space. During the mutation phase, the states of existing solutions are converted without changing their corresponding fitness values. After the $merge$ phase, $selection$ $for$ $replacement$ and $elitist$ $selection$ phases take place in order to guarantee the survival of the best individuals. The $elitist$ $selection$ stage is done in the whole population according to the fitness values of each state. Finally, a $split$ phase is necessary to disconnect all members of the whole population and reorder them in such a way that each homogeneous group reconstitutes a separate state. The $split$ and $merge$ cycle continues after each generation until the SEA obtains the ultimate solution or until a definite number of iterations is ``absorbed'' \cite{r20}. The main principle of the SEA is to choose the good state according to the fitness values of the actual solutions and not directly according to their states using a classical selection operator. Eventually, the SEA favours the coding whose the solutions have a best average fitness, and the choice of the coding depends on the evolvability of that coding just after the modification of representation. In another terms, it depends on the evolvability of the coding conversion operator which can be resumed as its capacity to promote and support the crossover and mutation operators to build new promising solutions from the old ones. Consequently, the conception of coding conversion operators that can lead to a scaled average fitness and an assorted evolvability would be a good approach to attack and dissolve the EAs problem caused by the representation issues. \\
On the other side, redundant representations are increasingly being applied in evolutionary computation and seem to affect positively the performance of genetic and evolutionary algorithms \cite{r27}. They use a higher number of alleles for encoding phenotypic information in the genotype than is essentially to construct the phenotype. This is the reason why we preface the use of block composites into the binary encoding which proves to have the features and properties of maintaining scaled genotypes and phenotypes (cf. Section \ref{secBBC}). In this paper, we expose an already-evoked structure of binary representation tagged as binary block coding (BBC) and founded on the concept of bitstrings decomposed into a definite number of blocks each having a fixed length \cite{r27}. The previous work in \cite{r27} discussed how the synonymy of a representation influences the genetic search. Then, it developed a population sizing model for synonymously redundant representations based on the assumption that a representation affects the initial supply. Our present study will focus on the framework of coupling more than one representation in one algorithm. Eventually, this paper will be centralized on the concept of proposing diverse ways of coding conversion and on the matter of how and when to apply these conversion operators which allow to change the representation of solutions in the population from one coding to another. In this intention, two different ways of changing the representation are evolved in this paper. The first one tends to increase the number of zeros ``$0s$'' in the binary solution by the fact that each bit will be encoded as a sub-solution composed by a block of binary bits having the maximum number of ``$0s$''. The second way tends to increase the number of ones ``$1s$'' in the binary solution by the fact that each bit in the solution will be encoded as a sub-solution composed by a block of binary bits having the maximum number of ``$1s$''. Thus, the conversion operators are different but the representations used to encode individuals in the population are identical by the fact that they have the same search space, the same neighborhood structure and the same fitness values for individuals having equivalent solutions. \\
The experiments are performed in intention to prove that even if the representations used to encode the solutions are identical, the manner of modifying the representation from one coding to another in the algorithm is of importance and is useful in the same way of selecting the best representation. This paper contains four main sections. In Section \ref{secBBC}, BBC is described in details. The entire set of experiments is exposed in Section \ref{secExp}. Section \ref{secGCC} presents general comments and concluding remarks. Finally, Section \ref{secFD} summarizes some further works.

\section{BBC and Conversion}
\label{secBBC}
The binary block coding and various coding conversion operators are evolved in this paper for the purpose of bringing some form of order into the disturbed situation caused by the influence of representation on the performance of EAs. The required specifications are outlined in the following subsections.

\subsection{Binary Block Coding}
\begin{figure}
\begin{center}
\begin{tabular}{@{}c@{\hspace{1mm}}c@{}}
\includegraphics[width=42mm,height=27mm]{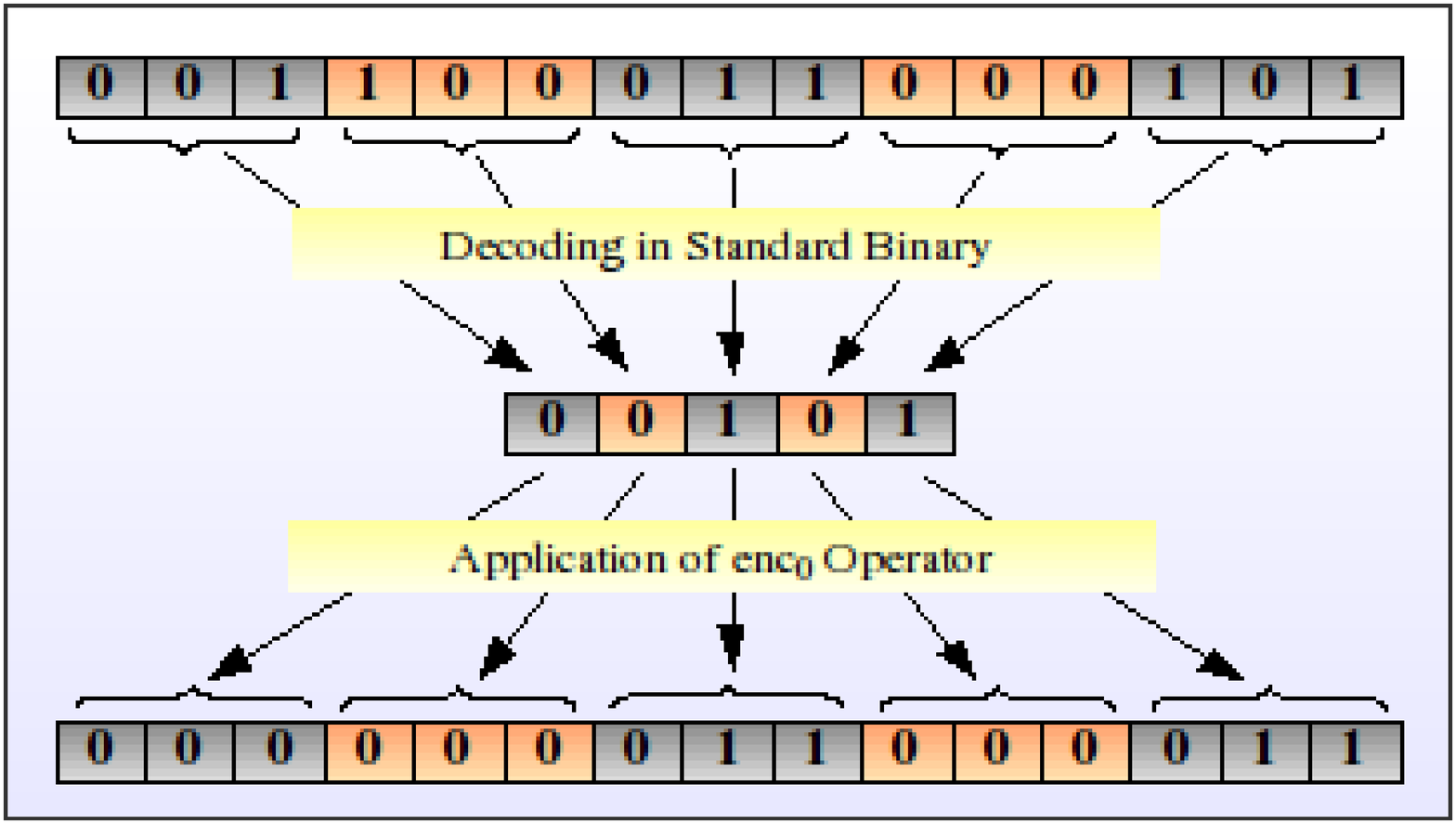} & \includegraphics[width=42mm,height=27mm]{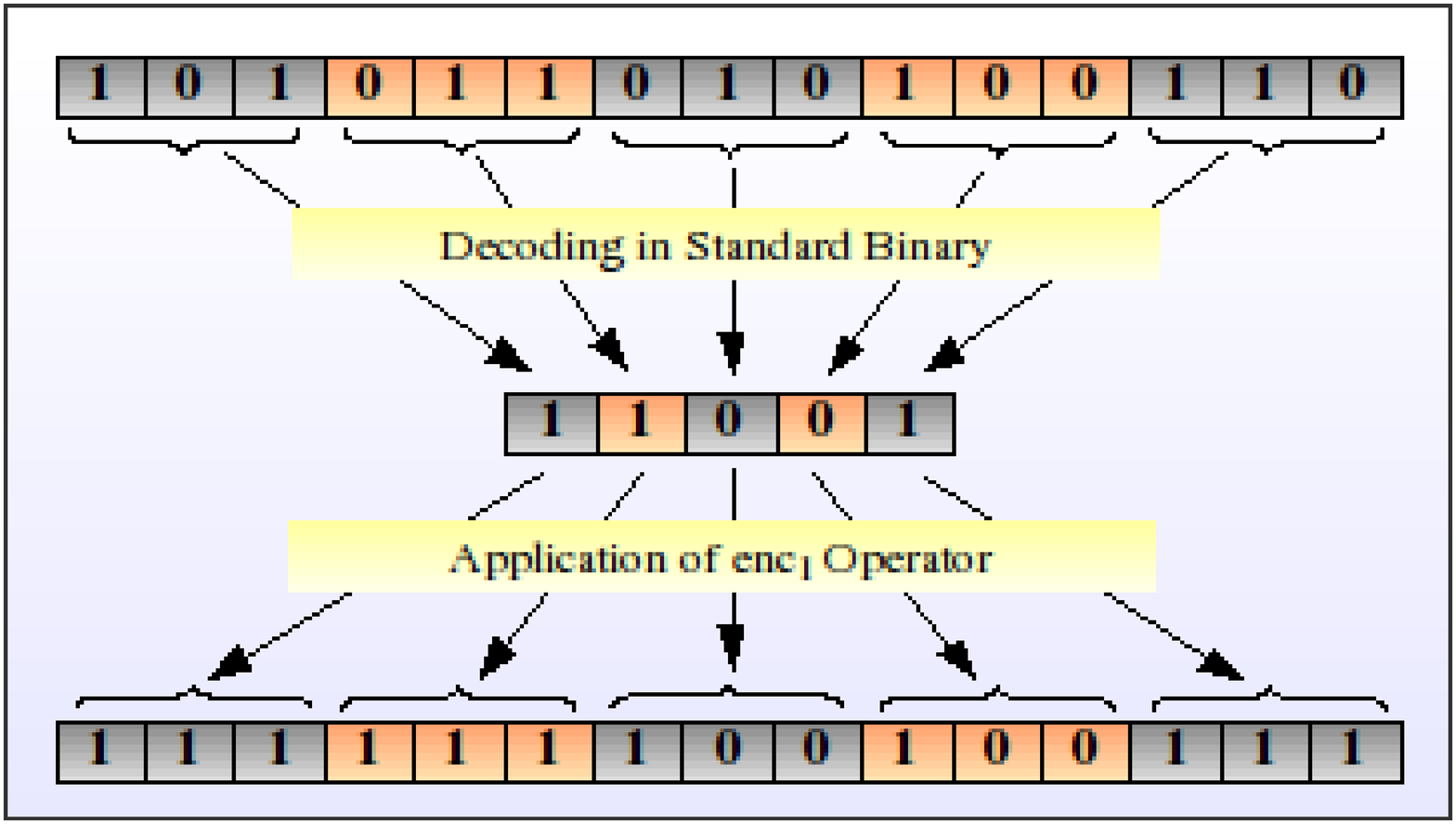}
\end{tabular}
\caption{For two given bitstrings $w$ and $w^\prime$ with a blocks number equal to $5$ and a block size equal to $3$ for both of them, we show the decoding in standard binary and then the application of $enc_0$ respectively of $enc_1$ operators.}
\label{figBBC}
\end{center}
\end{figure}

We present the binary block coding scheme, an existing binary representation which is based on the binary block constitution \cite{r27}. BBC is the set of all possible solutions $\{0,1\}^{nk}$ where $n$ is the blocks number and $k$ is the block size. Suppose that we have a bitstring $w$ which is encoded with BBC. $w$ will be composed of a set of binary blocks $w_j$ where $j$ $\in$ $[0,n-1]$ and each $w_j$ is of length $k$ (cf. Figure \ref{figBBC}). The decoding of $w$ to the standard binary returns a bitstring $x$ of length $n$. This procedure can be defined by the binary voting mapping. Each block in $w$ will be replaced by one bit in $x$. The value of each bit in $x$ is determinded by the ``voting to the majority of the values'' in the corresponding block. Generally, a specific binary encoded optimization task requires a binary representation that correlates to its fitness function structure. In this intention and since a bit value can be set equal to ``$0$'' or ``$1$'', we state two variants of changing the binary representation from one coding to another. The first is assigned to maximize the number of ``$0s$'' in the bitstring by transforming each bit in the string to a block of binary substring containing the largest possible number of ``$0s$'' (cf. Figure \ref{figBBC}). The second variant is assigned to maximize the number of ``$1s$'' in the bitstring by transforming each bit in the string to a block of binary substring containing the largest possible number of ``$1s$'' (cf. Figure \ref{figBBC}). BBC is considered to introduce a form of redundancy to the chromosome codification and is itself an infinite group of binary coding schemes by just varying the block size, and the standard binary coding is the basic element of this group with a block size equal to $1$.

\subsection{BBC Encoding Operators}
Suppose that we have a bitstring $x$ of length $n$ whose we want to encode in BBC with a block size equal to $k$ generating as well a new bitstring $w$. As it has been mentioned above, two encoding operators are available to change the representation from standard binary coding to BBC. The first, $enc_0$, maximizes the number of ``$0s$'' and the second, $enc_1$, maximizes the number of ``$1s$'' in the bitstring. So $\forall$ $i$ $\in$ $\{0,1\}$, $enc_i$ operator can be defined as follows:
$$
enc_i: \{0,1\}^n \rightarrow \{0,1\}^{nk}
$$
$$
enc_i(x) = w = w_0 w_1 ... w_{n-1}
$$
where $\forall$ $j$ $\in$ $[0,n-1]$,
$$
w_j =
\begin{cases}
i^k & \text{if } x_j = i \\
i^{\frac{k-1}{2}} \bar{i}^{\frac{k+1}{2}} & \text{if } x_j = \bar{i}
\end{cases}
$$
where $\bar{i}$ is the bitwise complement of $i$. \\
Two demonstrative examples are given in Figure \ref{figBBC}.

\subsection{BBC Decoding Operator}
Suppose that we have a bitstring $w$ composed of $n$ blocks each having a size equal to $k$. If we want to decode $w$ in standard binary, a new bitstring $x$ will be generated using the decoding operator $dec$. The decoding procedure from BBC to standard binary coding is based on a predefined function called $maj$ used to evaluate each $w_j$ in $w$ where $j$ $\in$ $[0,n-1]$. $maj$ routine is specified by the ``voting to the majority of the values'' in the bitstring and it can be outlined as follows:
$$
maj: \{0,1\}^k \rightarrow \{0,1\}
$$
$$
maj(u) =
\begin{cases}
0 & \text{if } |u|_0 > |u|_1 \\
1 & \text{otherwise}
\end{cases}
$$
where $|u|_0$ respectively $|u|_1$ represents the number of ``$0s$'' respectively of ``$1s$'' in $u$. Then, $dec$ operator can be defined as follows:
$$
dec: \{0,1\}^{nk} \rightarrow \{0,1\}^n
$$
$$
dec(w) = x = x_0 x_1 ... x_{n-1}
$$
where $\forall$ $j$ $\in$ $[0,n-1]$,
$$
x_j = maj(w_j)
$$
Two demonstrative examples are given in Figure \ref{figBBC}.

\subsection{BBC Conversion Operators}
There exist several ways to change the representation of individuals in the population from BBC to BBC. We state below a brief list of two BBC conversion operators. They are transmutation utilities that change the state, here the representation, of a solution without changing the fitness value of that solution. So, we have $\forall$ $j$ $\in$ $S$, $\forall$ $x$ $\in$ $\Omega$, $f(x)$ $=$ $f(conv_j(x))$ where $S$ is the state space, $\Omega$ is the search space, $f$ the fitness function, and $conv_j$ the conversion operator to state $j$. In our case, $conv_0$ corresponds to the operator that maximizes the number of ``$0s$'' and $conv_1$ corresponds to the operator that maximizes the number of ``$1s$'' in the bitstring. Suppose that we have a bitstring $w$ encoded with BBC and composed of $n$ blocks each having a size equal to $k$. If we want to change the representation of $w$ to BBC in a form of redundancy that increases the number of ``$is$'' in the bitstring with the same block size where $i$ $\in$ $\{0,1\}$, a new bitstring $w^\prime$ will be generated following two main steps. The first belongs to the decoding of $w$ in standard binary producing as well a new bitstring $x$ of length $n$. The second step belongs to the encoding of $x$ in BBC by applying $enc_i$ operator poducing as well a new bitstring $w^\prime$ of length $nk$. Therefore $\forall$ $i$ $\in$ $\{0,1\}$, $conv_i$ operator can be defined as follows:
$$
conv_i: \{0,1\}^{nk} \rightarrow \{0,1\}^{nk}
$$
$$
conv_i(w) = w^\prime = enc_i(dec(w))
$$
Two demonstrative examples are given in Figure \ref{figBBC}.

\subsubsection{Role and Importance}
Some classes of optimization problems can take advantage from the coexistence and the application of the two BBC conversion operators, $conv_0$ and $conv_1$, in one algorithm. A dual coding strategy based on these two variants and developed genuinely in an EA serves to make the representation of solutions more adaptive and well-matched to a problem's fitness function. Likewise, this approach can make EAs advantageously explore undiscovered areas of the search space. If we introduce the notion of state to be defined according to the representation in a SEA, then that SEA can be the appropriate algorithm that integrates an adaptive approach for the representation in which the genotype encoding is altered dynamically by the fact that a state mutation will be equivalent to a coding conversion. Therefore, the modification of the representation of arbitrary solutions to a form of BBC using $conv_0$ or $conv_1$ tries to make an equilibrum in the number of bits with ``$0$'' and ``$1$'' while a classic binary representation sometimes makes bias towards the bits with ``$0$'' or ``$1$''. For example, if the ultimate solution of an optimization problem contains a number of ``$0s$'' more than the number of ``$1s$'' in the string then the BBC coding alternation ``tour'' performed in a SEA while applying $conv_0$ and then $conv_1$ to random solutions during the search may be helpful in increasing the number of bits with ``$0$'' and then can lead, iteration after iteration, to discover and locate the global optimum. The role of BBC conversion operators can be seen as intermediators between the standard binary coding and the problem structure, and those mediators serve to well explore new regions in the search space. The importance of those operators lies on the concept that specifies them as adjustors which attempt to correct the erroneous bits in the string by replacing each probable false bit value by the true one, the matter which can be seen and interpreted indirectly as the constructors of the meaningful building blocks. Since in a binary coding, ``$0$'' is the bitwise complement of ``$1$'' and inversely ``$1$'' is the bitwise complement of ``$0$'', so $conv_0$ can be translated as the complementary conversion operator of $conv_1$ and reciprocally $conv_1$ can be translated as the complementary conversion operator of $conv_0$. In this aim, we must notice that the value of BBC resides in using $conv_0$ and $conv_1$ operators simultaneously in one method that let them interact and interchange data bits to finally assisst in creating and not in destroying the substantive building blocks.

\subsubsection{Evolvability}
The evolvability of a coding conversion operator is defined as the phenotypic effects that can be produced after the change of the representation of solutions using that operator. In another terms, it is the ability of that operator to affect and serve the genetic operators, crossover and mutation, to develop new promising solutions from the old ones during the reproduction phase. Particularly, the evolvability of a coding conversion operator deeply depends on the problem structure and the shape of the optimum. Suppose that we have to optimize a problem where the global optimum contains a number of ones ``$1$'' greater than that of zeros ``$0$''. Then, $conv_1$ operator will be more favored regarding its concern in maximizing the number of ones ``$1$'' in the bitstring. Consequently, the chance to produce new promising solutions after the application of $conv_1$ will be greater than that after the application of $conv_0$, and hence the evolvability of $conv_1$ will be greater than that of $conv_0$. Two experimental tests were performed in sections \ref{bbcanl} and \ref{secGCC} to study and compare the evolvability of $conv_0$ and $conv_1$ operators.

\section{Experiments}
\label{secExp}
We have prepared a set of experiments to test and analyze some of the main features of BBC conversion operators in use with a SEA, and to show the importance of changing the representation of solutions during the search process.

\begin{table}
\begin{center}
\caption{Test Functions}
\label{tabFunc}
\begin{tabular}{lll}
\hline
{\bf Reference} & {\bf Name} & {\bf Definition} \\
\hline
P1 & ONEMAX & $f_1(s) = |s|_1$ \\ \\
P2 & NEEDLE & $f_2(s) = \begin{cases} l & \text{if } |s|_1 = l \\ 1 & \text{otherwise} \end{cases}$ \\ \\
P3 & ONOFF & $f_3(s) = HDO(s)$ \\ \\
P4 & ALTERNATION & $f_4(s) = ND(s)$ \\
\hline
\end{tabular}
\end{center}
\end{table}

\subsection{Test Functions}
\label{secTestFunc}
\begin{figure}
\begin{center}
\begin{tabular}{@{}c@{}c@{}}
\includegraphics[width=42mm,height=30mm]{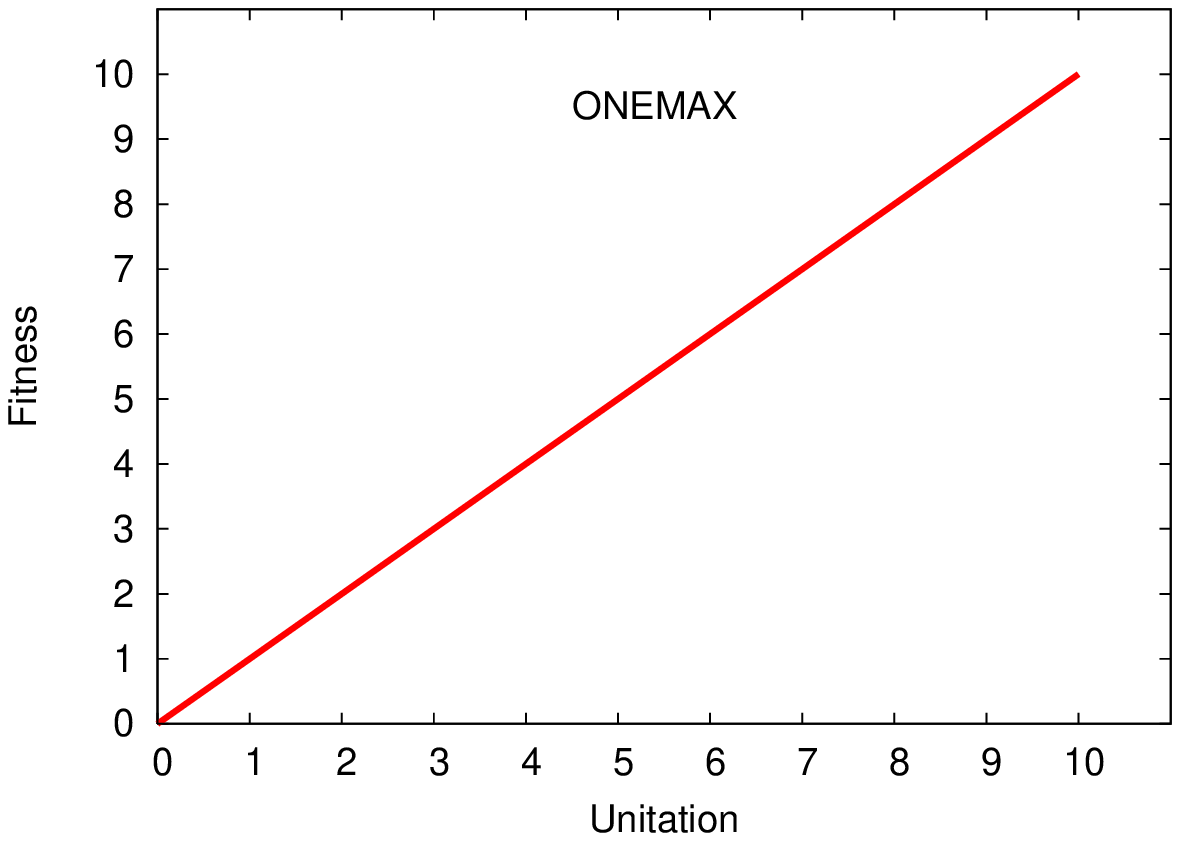} & \includegraphics[width=42mm,height=30mm]{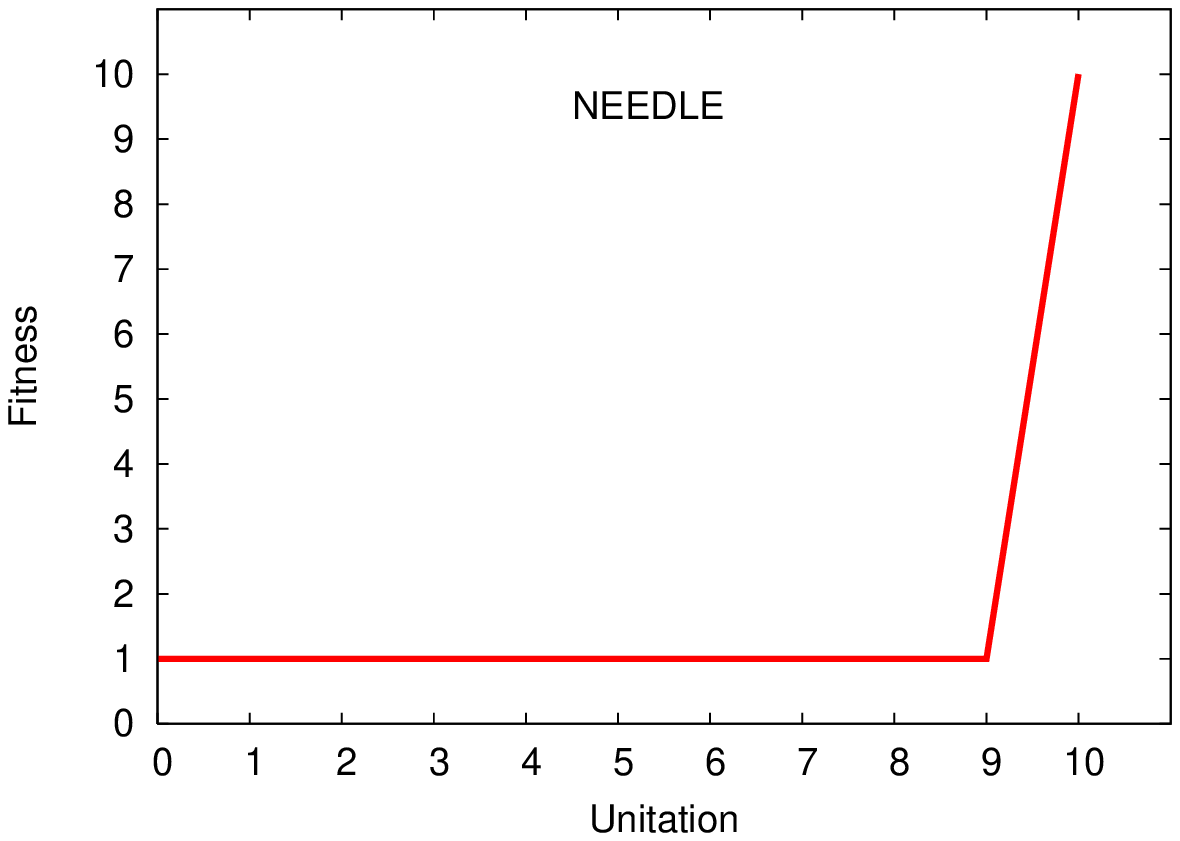} \\
ONEMAX & NEEDLE
\end{tabular}
\caption{Graphical representations of unitation functions.}
\label{figFunc}
\end{center}
\end{figure}

\begin{figure}
\begin{center}
\begin{tabular}{@{}c@{}c@{}}
\includegraphics[width=42mm,height=30mm]{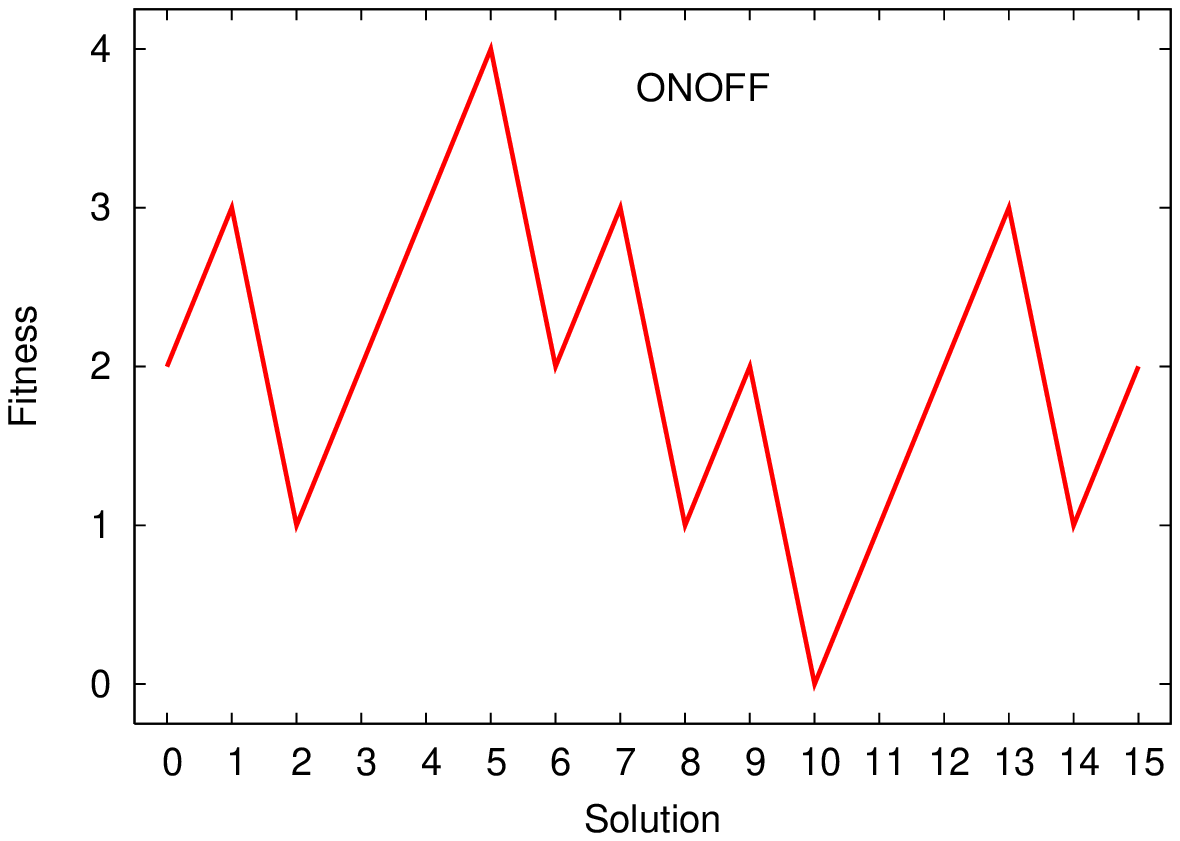} & \includegraphics[width=42mm,height=30mm]{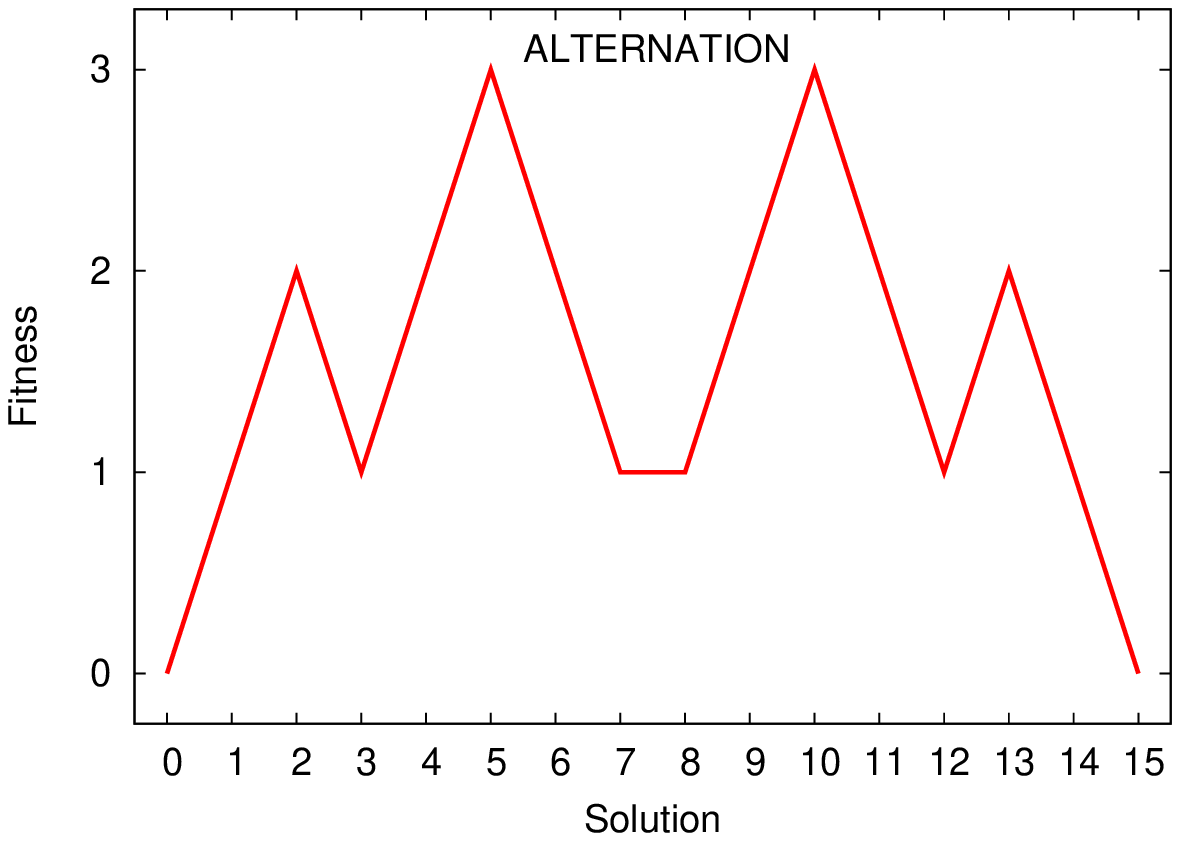} \\
ONOFF & ALTERNATION
\end{tabular}
\caption{Examples of the graphical representation of the ONOFF and ALTERNATION functions. For a binary solution of length equal to $4$, we compute the fitness values corresponding respectively to the $16$ ($2^4$) possible solutions. The x-axis represents the real-value of each standard binary solution. The y-axis represents the fitness of solutions.}
\label{figFunc2}
\end{center}
\end{figure}

To test the performance of optimization algorithms, standard test problems should be used. We mainly consider a set of four binary encoded optimization functions. \\
The first one is P1 and is the classical ONEMAX problem. It belongs to the unitation class of fitness functions. Unitation functions are fitness functions where the fitness is a function of the count of ``$1s$'' in a solution $x\in\{1,0\}^l$, where $l$ is the length of the solution. All fitness values are non-negative: $u:\{0,1\}^l \rightarrow \R^+$. The first two fitness functions given in Table \ref{tabFunc} and pictured in Figure \ref{figFunc} are two examples of unitation functions. They are respectively called ONEMAX and NEEDLE, and have been theoretically studied for fixed parameter simple GAs by Rowe \cite{r22}, Wright \cite{r23} and Richter et al. \cite{r24}. The ONEMAX fitness function has been called the ``fruit fly'' of GA research \cite{r25}. It is a maximization problem that countes the number of ``$1s$'' in the string. P1 is a neutral linear function with one global optimum, an all ``$1s$'' string. \\
As well, we have expanded our observations to test the second function P2. It is the NEEDLE problem which also belongs to the unitation class of fitness functions. P2 has one global optimum, an all ``$1s$'' string, and is reasoned to be a difficult optimization task for the classic GA to work out. NEEDLE is a maximization linear problem and can serve to study the properties of the SEA and show the importance of changing the representation. \\
On the other side, we have applied our tests on the ONOFF problem P3. We define the ONOFF problem as a fitness function where the global optimum is a finite binary sequence of the form $1010...10$ and the fitness is the regular Hamming distance of a solution $x\in\{1,0\}^l$ to the global optimum, where $l$ is the length of $x$. ONOFF is a typical minimization problem. All fitness values are non-negative: $u:\{0,1\}^l \rightarrow \R^+$, and the fitness value of the global optimum corresponds to a value of $0$ for any length of the solution. Each bit of value $1$ in the binary string of the global optimum represents the ON label and each bit of value $0$ represents the OFF label. An illustrative example of the ONOFF function is pictured in Figure \ref{figFunc2} for a length of the binary solution equal to $4$. This function should advantageously confirm our assertions about changing the representation because we consider that the genuine solution of the form $1010...10$ will be a really challenging task for BBC conversion operators. Consequently, $conv_0$ and $conv_1$ operators should have the equal opportunities to be applied during the optimization task regarding the global optimum that contains an equal and consecutive number of ``$0s$'' and ``$1s$''. \\
Likewise, the experiments are extended to include the ALTERNATION problem P4. This function counts the number of dicontinuities between consecutive bits in the bitstring \cite{r44}. It is considered as a hard maximization problem for a simple GA to solve. It depends on the total number of sequences $10$ or $01$ in a string and not on the positions of the alternations. So, it is defined on the binomial distribution of the space induced by alternations. As a consequence, ALTERNATION function has the following properties: 1) All the points with the same number of alternations have the same fitness value. 2) Symmetry with respect to bit value, that is $f(x)=f(\bar{x})$, where $x\in\{1,0\}^l$ is a bitstring of length $l$, and $\bar{x}$ is its bitwise complement. 3) According to the above property, the fitness Hamming distance correlation coefficient is equal to zero. An illustrative example of the ALTERNATION function is pictured in Figure \ref{figFunc2} for a length of the binary solution equal to $4$. This problem provides an interesting tool to analyze and report the dimensions of $enc_0$ and $enc_1$ operators by the fact that it features two global optimum regarding its symmetry characteristic. The first is of the shape $1010...10$ and the second one is of the shape $0101...01$. Contrarily to the first three problems, P4 is a non-linear problem where a form of epistasis is contained in the structure of the solution and the bits are tightly linked each to other. The chances to apply $conv_0$ and $conv_1$ operators must be equivalent for the EA to succeed. \\
The definitions of all these problems are summarized in Table \ref{tabFunc} where $l$ is the length of the solution $s$, $|s|_1$ is the number of ``$1s$'' in $s$, $HDO(s)$ is the Hamming distance of $s$ to the global optimum, and $ND(s)$ is the count of dicontinuities between consecutive bits in $s$. In order to compute the fitness value $f^\prime$ of a given solution $w$ which is encoded by BBC where $f^\prime:\{0,1\}^{nk} \rightarrow \R^+$, first we decode $w$ in standard binary generating as well a new bitstring $x$. And then, the fitness value $f$ of $x$ is taken equal to the corresponding function value which is calculated according to the function expression given in Table \ref{tabFunc} where $f:\{0,1\}^n \rightarrow \R^+$. And so, we obtain the following equality: $f^\prime = f \circ dec$.

\subsection{SEA Parameters}
Since a SEA itself has several options in terms of its implementation, it is necessary to denote the parameter choices used in this paper. First, the number of states was set to $2$, each state being represented by a simple GA ({\small S}GA). The representations of solutions applied in all states are identical and we used BBC for encoding the solutions. This choice was explicit in intention to integrate a dual coding strategy which may help in locating the ultimate solution while changing the representation of different random solutions from one coding to another using BBC conversion operators. Thus, we have two states symbolized by two {\small S}GAs which are similar in everything and each component. And so, an instance of the SEA is implemented and is denoted 2-SEA where the representation is directly linked to the algorithm and not to the individual in the population. Each {\small S}GA is executed for one simple iteration before the $merge$ phase takes place in the algorithm life-cycle. Next, the follow-up parameter is $pMutState$ for state mutation rate. In our case and since the representation is directly linked to the algorithm, a state mutation means that the representation of individuals in that algorithm is changed to another representation. Afterwards, we will refer to $pMutState$ by conversion rate. $pMutState$ parameter could be easily modified to provide conversion of arbitrary solutions from one coding scheme to another without affecting the results dramatically during the search process. The value of this parameter is fixed using the experiments described later in Section \ref{bbcanl}. As well, 2-SEA has another particular parameter: $k$ for the block size. The value used for that parameter was chosen as a result of prior experimentation reported subsequently in Section \ref{bbcanl}. The best parameter settings between those tested for all objective functions are given in Table \ref{tabParam}.

\subsection{General Parameter Values}
In order to create a fair tableau for comparison of {\small S}GA with 2-SEA, the parameters shared between these two algorithms were kept the same. Since 2-SEA is composed of two parallel {\small S}GAs, the classic GA and 2-SEA were run with the parameters recommended by Goldberg (Goldberg 1989) (cf. Table \ref{tabGPV}). In general, the set of all used parameters and their respective attributes are shown in Table \ref{tabParam} with: $maxGen$ for maximum number of generations before STOP, $popSize$ for population size, $vecSize = nk$ for genotype size, $tSize$ for tournament selection size, $pCross$ for crossover rate, $pMut$ for mutation rate, and $pMutPerBit$ for bit-flip mutation rate. This tableau was employed for the four test problems. We have to mention that for the first three test problems the population size was set equal to $100$ and for the last problem this parameter value was set equal to $10$ which reflects the fact that the ALTERNATION function requires more exploitation than exploration due to the deceptive attractor which is at mid-distance from the global optimum. This choice is well verified and is totally compatible with the choice of a low $pMutPerBit$ value for the ALTERNATION function which enables the algorithm to discover recursively and regularly good directions in the search interval.

\begin{table}
\begin{center}
\caption{General Parameter Values}
\label{tabGPV}
\begin{tabular}{|@{\hspace{1.8mm}}l@{\hspace{1.75mm}}|@{\hspace{1.8mm}}l@{\hspace{1.75mm}}|}
\hline
{\bf Parameters} & {\bf Attributes} \\
\hline
$Pseudorandom$ $generator$ & Uniform Generator \\
\hline
$Selection$ $mechanism$ & Tournament Selection \\
\hline
$Crossover$ $mechanism$ & 1-Point Crossover \\
\hline
$Mutation$ $mechanism$ & Bit-Flip Mutation \\
\hline
$Replacement$ $model$ $1$ & Generational Replacement \\
\hline
$Replacement$ $model$ $2$ & Elitism Replacement \\
\hline
$Ending$ $criteria$ & Maximum Number of Iterations \\
\hline
\end{tabular}
\end{center}
\end{table}

\begin{table}
\begin{center}
\caption{Best Parameter Settings}
\label{tabParam}
\begin{tabular}{|l|@{\hspace{4.2mm}}c@{\hspace{4.2mm}}|@{\hspace{4.2mm}}c@{\hspace{4.2mm}}|@{\hspace{4.2mm}}c@{\hspace{4.2mm}}|@{\hspace{4.2mm}}c@{\hspace{4.2mm}}|}
\hline
{\bf Parameters} & {\bf P1} & {\bf P2} & {\bf P3} & {\bf P4} \\
\hline
$maxGen$ & $3000$ & $3000$ & $3000$ & $30000$ \\
\hline
$popSize$ & $100$ & $100$ & $100$ & $10$ \\
\hline
$vecSize$ & $1900$ & $1900$ & $300$ & $300$ \\
\hline
$tSize$ & $2$ & $2$ & $2$ & $2$ \\
\hline
$pCross$ & $0.6$ & $0.6$ & $0.6$ & $0.6$ \\
\hline
$pMut$ & $1.0$ & $1.0$ & $1.0$ & $1.0$ \\
\hline
$pMutPerBit$ & $0.9$ & $0.9$ & $0.05$ & $0.05$ \\
\hline
$pMutState$ & $1.0$ & $1.0$ & $0.85$ & $0.7$ \\
\hline
$k$ & $19$ & $19$ & $3$ & $3$ \\
\hline
\end{tabular}
\end{center}
\end{table}

\subsection{Experimental Results}
In the following two subsections, we introduce the experiments that have been performed for two different purposes. The first serves to analyze BBC conversion operators and to study the interaction and the dependency of the parameters of both BBC and 2-SEA. And the second purpose tries to test the importance of changing the representation and contributes in a comparison between the performance of 2-SEA and the classic GA.

\subsubsection{BBC Analysis}
\label{bbcanl}
\begin{figure}
\begin{center}
\begin{tabular}{@{}c@{\hspace{1mm}}c@{}}
\includegraphics[width=42mm,height=30mm]{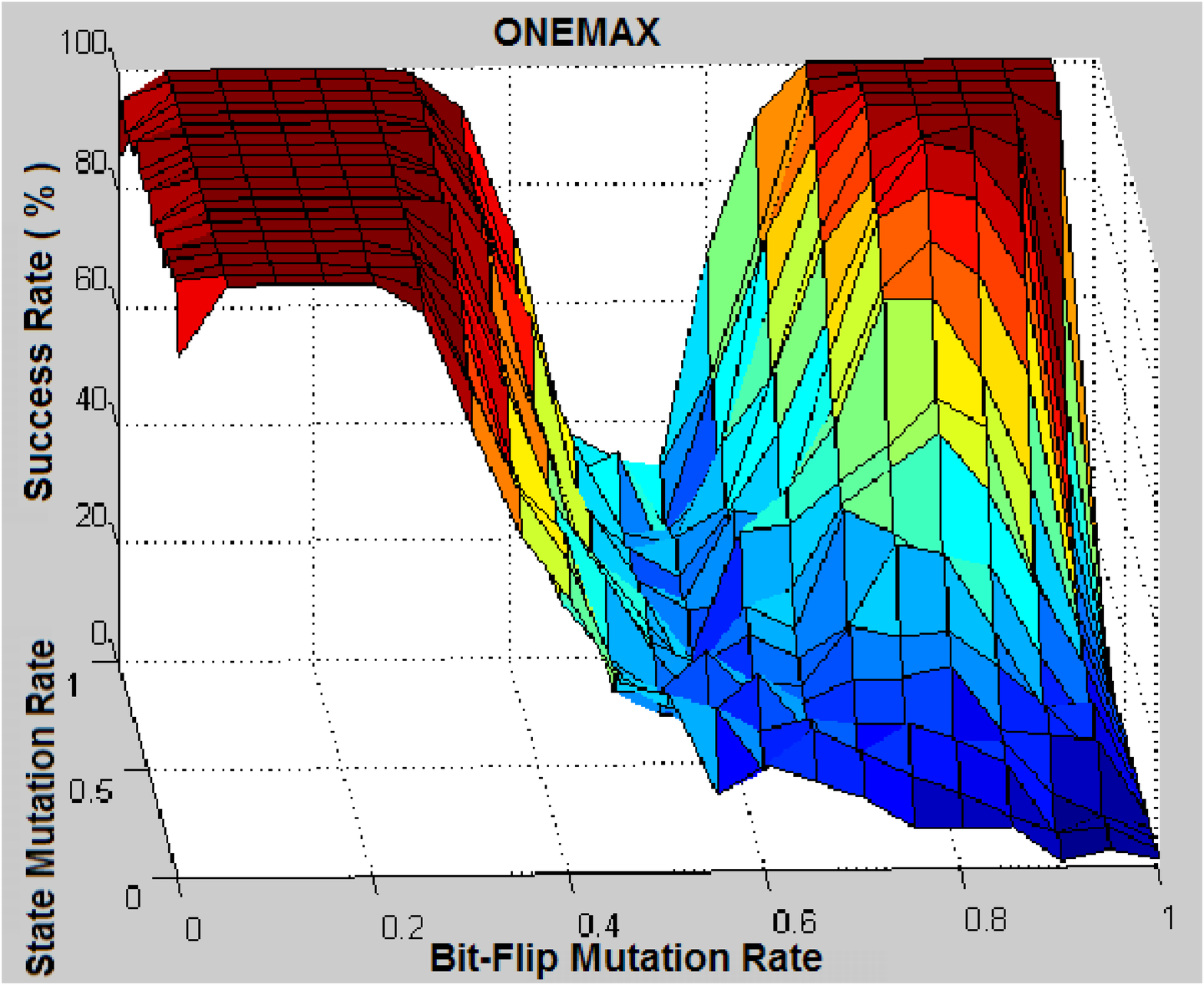} & \includegraphics[width=42mm,height=30mm]{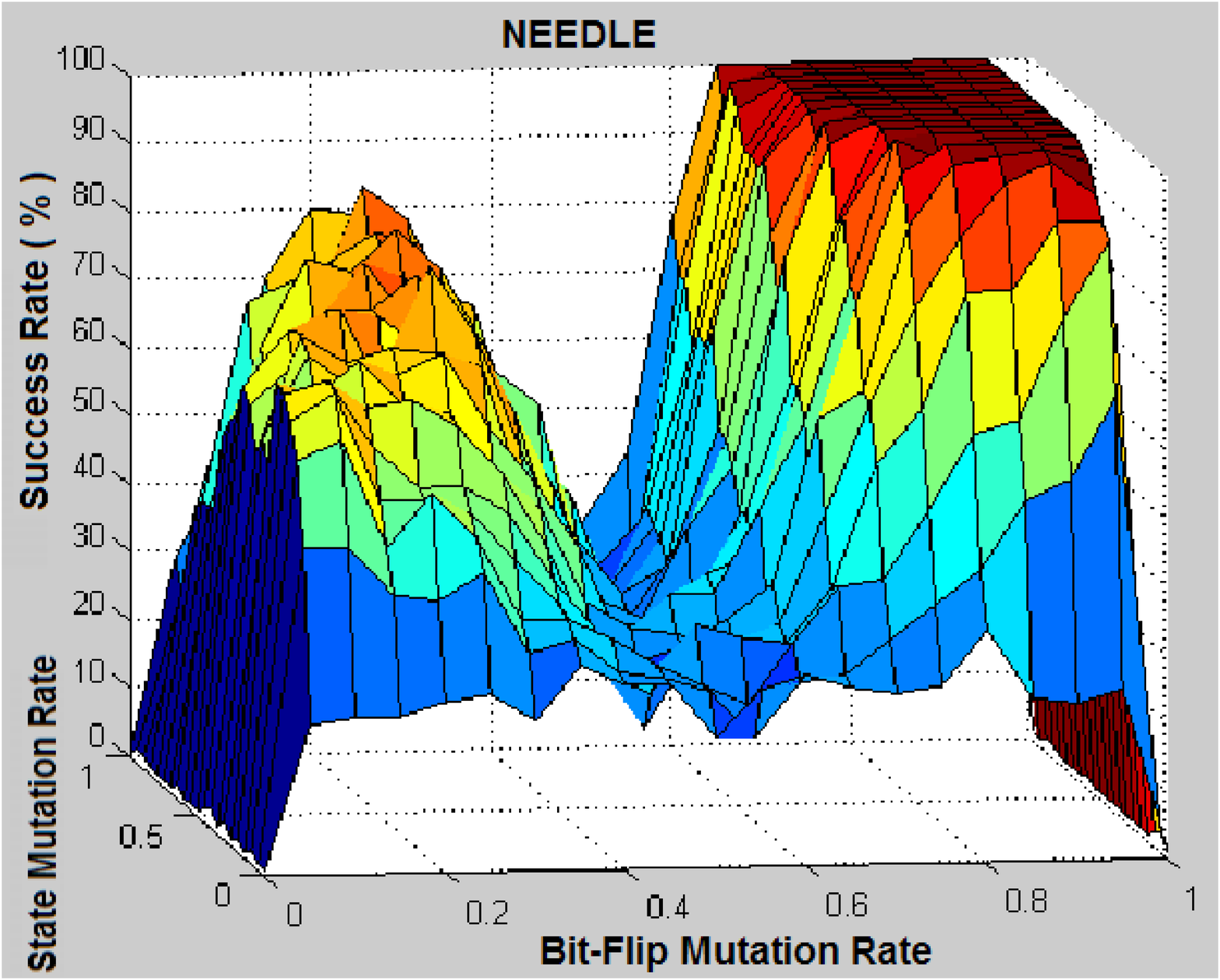} \\
\includegraphics[width=42mm,height=30mm]{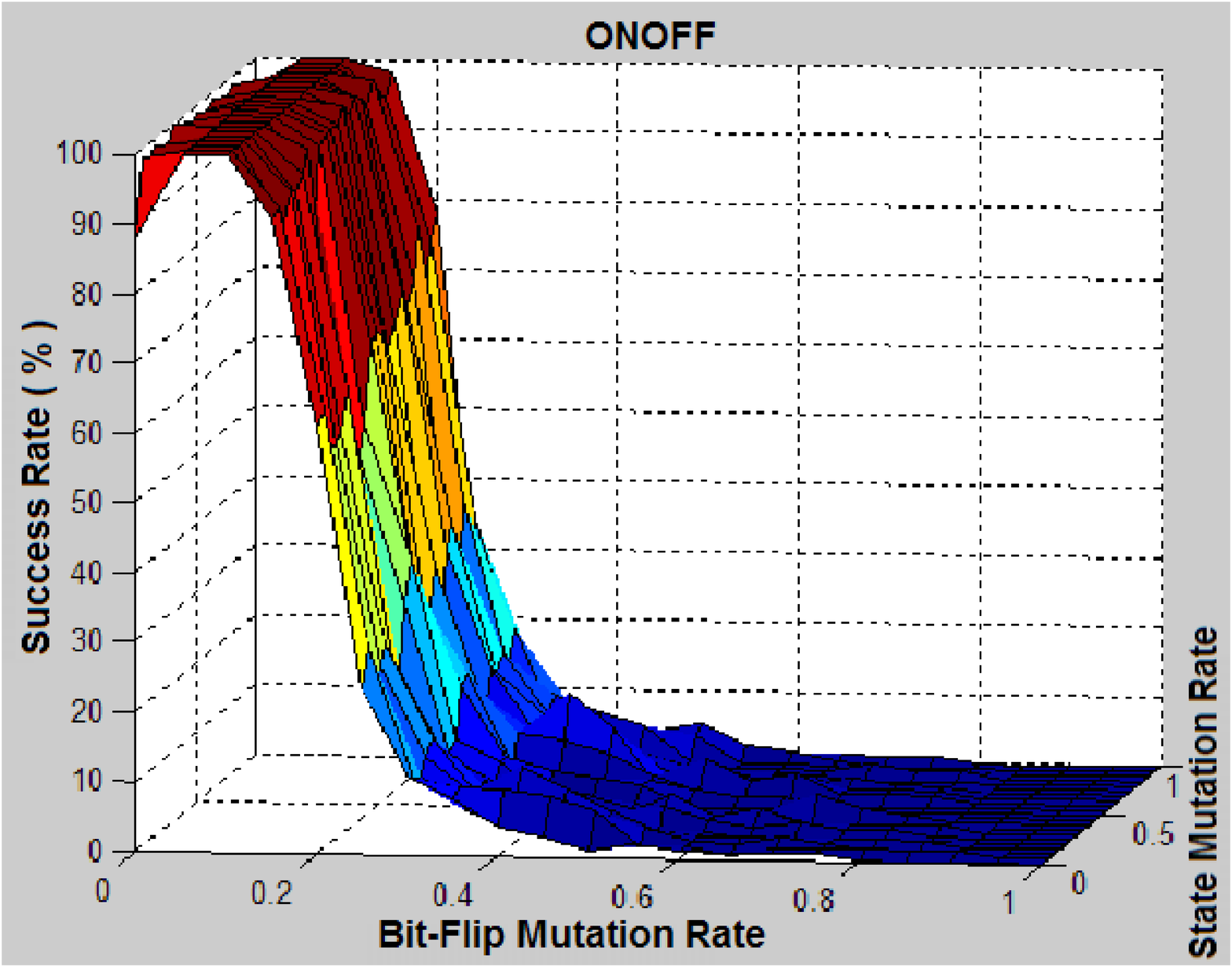} & \includegraphics[width=42mm,height=30mm]{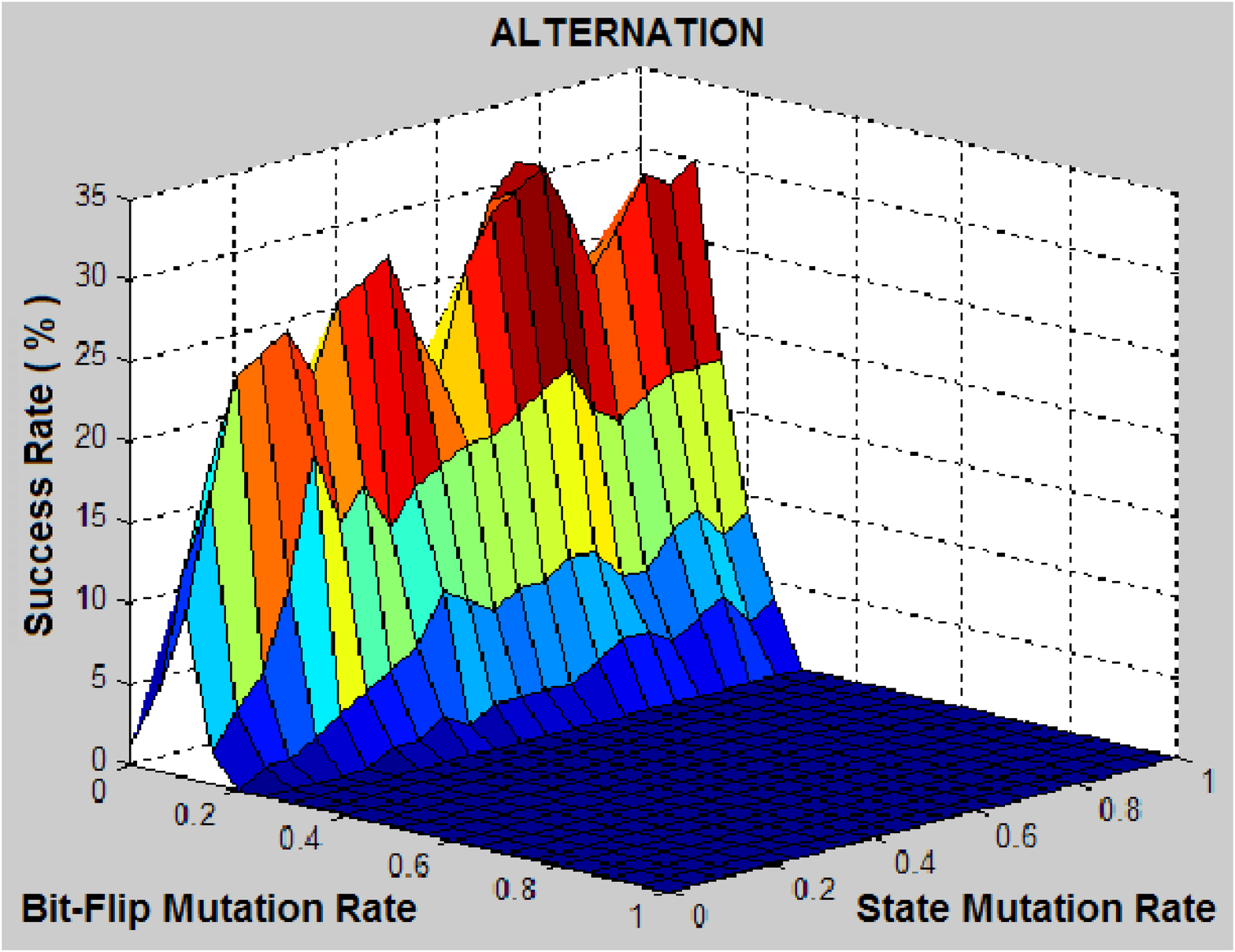} \\
\end{tabular}
\caption{Study of the success rate fluctuations relatively to the variations of the conversion rate and the bit-flip mutation rate. $pMutState$ and $pMutPerBit$ values varied from $0.0$ to $1.0$ with a step of $0.05$ for all test functions. As a result, the obtained success rate values varied from $0\%$ to $100\%$.}
\label{fig-er-functions}
\end{center}
\end{figure}

\begin{figure}
\begin{center}
\begin{tabular}{@{}c@{}c@{}}
\includegraphics[width=42mm,height=30mm]{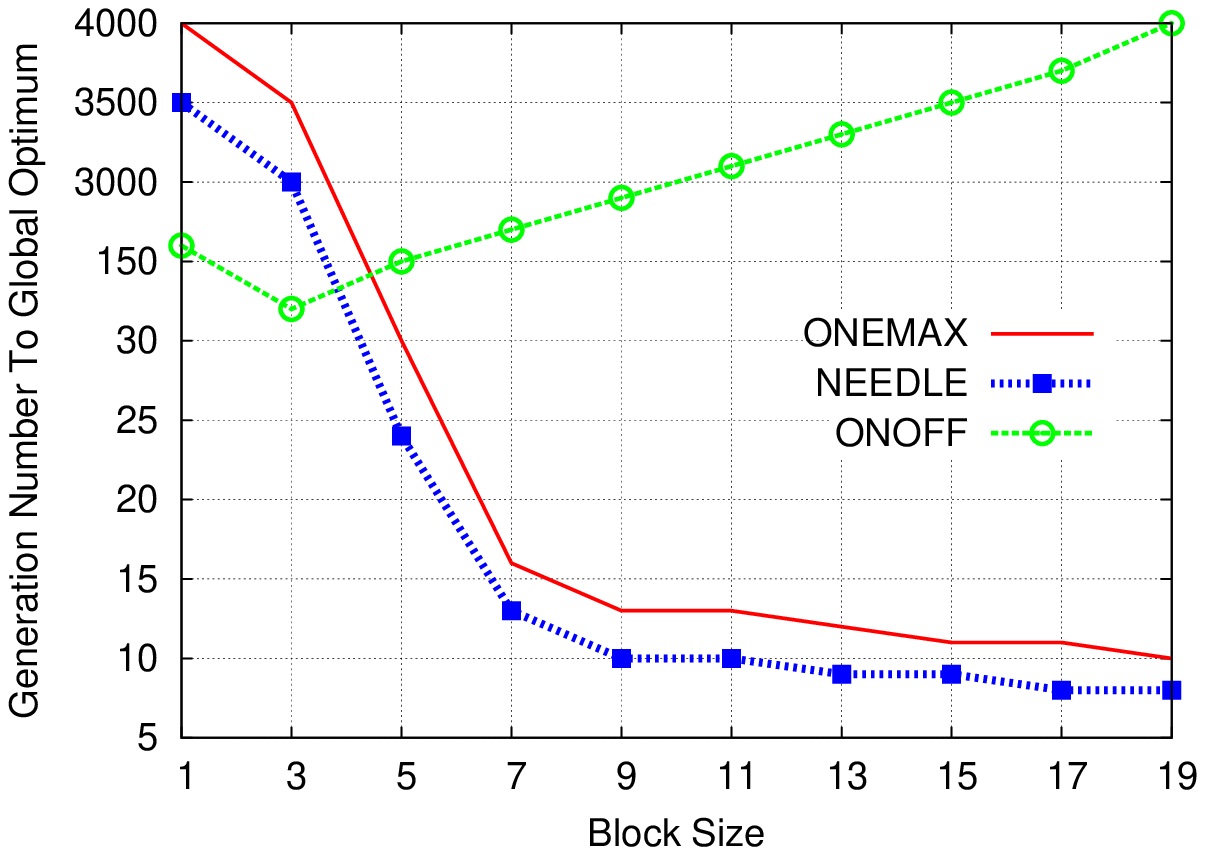} & \includegraphics[width=42mm,height=30mm]{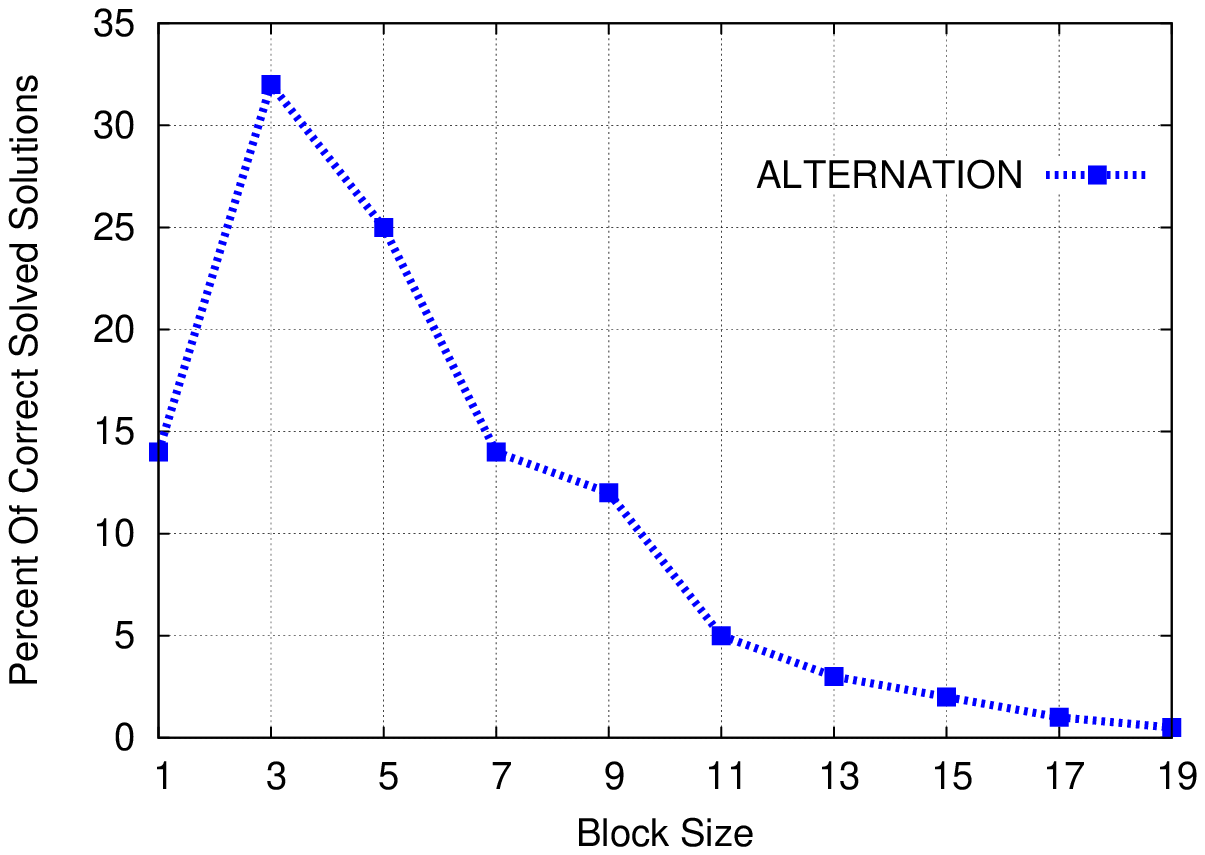} \\
\end{tabular}
\caption{For the best values of $pMutState$ and $pMutPerBit$, we plot the number of iterations required to reach the global optimum relatively to the block size for P1, P2 and P3 problems. As well, we plot the percent of correct solved solutions relatively to the block size for P4 problem.}
\label{figBS}
\end{center}
\end{figure}

In this section, we present experiments designed to examine several aspects of BBC conversion operators. We would like to know how much the change of the representation using $conv_0$ and $conv_1$ could ``help'' and ``advance'' 2-SEA during the search. Besides, we would like to discover how the parameters of both BBC and 2-SEA interact each with other. \\
First, it is so essential to mention that $pMutState$ and $pMutPerBit$ parameters play an important role in 2-SEA operation, and their affected values are decisive in the final outcome. Precisely, $pMutState$ is responsible for the conversion of arbitrary individuals in the population from their initial representation to the other one. In our research, we are using $conv_0$ and $conv_1$ as two different conversion operators for the same search space defined by BBC. Consequently, each of these two operators has a different evolvability after the change of the representation. Since the evolvability of a coding conversion operator and with it $pMutState$ is incidental to the application of genetic operators and with it to the probability of flipping one bit in a bitstring, $pMutPerBit$, we will begin by exploring the relationship between $pMutState$ and $pMutPerBit$ parameters and the proportion of solutions solved correctly by 2-SEA, success rate in percent. In the first experiment, $pMutState$ and $pMutPerBit$ values changed within [$0.0:1.0$] interval with a step of $0.05$. This experiment was realized on each test problem for $100$ independent runs. Graphical representations of fitness variations relatively to $pMutState$ and $pMutPerBit$ were given in Figure \ref{fig-er-functions}. A simple reading of these figures shows that a large conversion rate is needed for all test functions in order for 2-SEA to produce positive results which reflects the great importance and utility of the change of the representation during the search. Besides, Figure \ref{fig-er-functions} indicates that a high bit-flip mutation rate is required for P1 and P2 problems, and a small bit-flip mutation rate is required for P3 and P4 problems so that 2-SEA can render important end results. As an elementary synthesis on these obtained results, we can say that P3 and P4 problems require a low-level of mutation effects regarding the ordered structure of their global optimums which necessitate a modest contribution of the genetic operators, especially the bit-flip mutation, to be able to rearrange and fix up each bit in its correct position in the bitstring. \\
On the other side, BBC has another key parameter: $k$ for the block size. The second experiment is performed to determine the value of that parameter for each test function. First, we have fixed the length of the standard binary genotype to a value of $n = 100$ which means that the number of blocks in the binary block genotype will be equal to $100$ and the fitness value of the global optimum will be equal to $100$ for unitation functions, $0$ for P3 problem, and $99$ for P4 problem. Likewise, we have fixed the values of $pMutState$ and $pMutPerBit$ parameters respectively to $1.0$ and $0.9$ for unitation functions. P3 and P4 problems have a $pMutState$ value equal to $0.85$ respectively $0.7$ and a $pMutPerBit$ value equal to $0.05$ for both of them. This test was realized on all objective functions for $100$ independent runs. Graphical records are displayed in Figure \ref{figBS} and show that a large block size is necessary for the unitation functions in order for 2-SEA to produce significant positive results in a minimum number of iterations. This fact can be explained as a consequence of that, for classical linear problems, an optimal evolvability of a coding conversion operator is related to a maximal length of a bitstring and next to a maximal or large block size. On the other side, P3 and P4 problems require a small block size to make 2-SEA competent to submit large-scale solutions by the concept of that a small $pMutPerBit$ value and with it a low-order evolvability can avoid a disruptive effect on the solution and as a result it can help in adjusting the structural form of individuals heuristically and progressively in a minor number of generations.

\subsubsection{Performance Comparison: 2-SEA vs. {\small S}GA}
Considering the stochastic nature of 2-SEA, we compute the average performance of $100$ independent runs of 2-SEA on each objective function. The global optimum being equal to $100$ for P1 and P2, $0$ for P3 and $99$ for P4, Table \ref{tabExpRes} shows the numerical results whereas Figure \ref{figExpRes} represents the graphical records of the experiments. For the two algorithms, {\small S}GA and 2-SEA, Table \ref{tabExpRes} displays two main records for each test function. The first is the success rate (SR$_\%$) measurement and is the percentage of the number of runs in which the algorithm succeeded in finding the global optimum. The second record is the generation number to optimum (GNTO) measurement and is the average of the number of iterations needed for the algorithm to attain the global optimum.

\begin{table}
\begin{center}
\caption{Experimental Results}
\label{tabExpRes}
\begin{tabular}{|l|l|c|c|}
\hline
{\bf Problem} & {\bf Measurement} & \multicolumn{2}{c|}{\bf Algorithm} \\
\cline{3-4}
 & & \hspace{5.5mm}{\bf {\small S}GA}\hspace{5.5mm} & \hspace{5.5mm}{\bf 2-SEA}\hspace{5.5mm} \\
\hline
\multirow{2}{*}{\bf P1} & {\bf SR $\%$} & $100$ & $100$ \\
\cline{2-4}
 & {\bf GNTO} & $128$ & $10$ \\
\hline
\multirow{2}{*}{\bf P2} & {\bf SR $\%$} & $3$ & $100$ \\
\cline{2-4}
 & {\bf GNTO} & $3000+$ & $8$ \\
\hline
\multirow{2}{*}{\bf P3} & {\bf SR $\%$} & $100$ & $100$ \\
\cline{2-4}
 & {\bf GNTO} & $579$ & $84$ \\
\hline
\multirow{2}{*}{\bf P4} & {\bf SR $\%$} & $4$ & $32$ \\
\cline{2-4}
 & {\bf GNTO} & $30000+$ & $30000+$ \\
\hline
\end{tabular}
\end{center}
\end{table}

\section{Discussion and Conclusion}
\label{secGCC}
\begin{figure}
\begin{center}
\begin{tabular}{c}
\includegraphics[width=80mm,height=50mm]{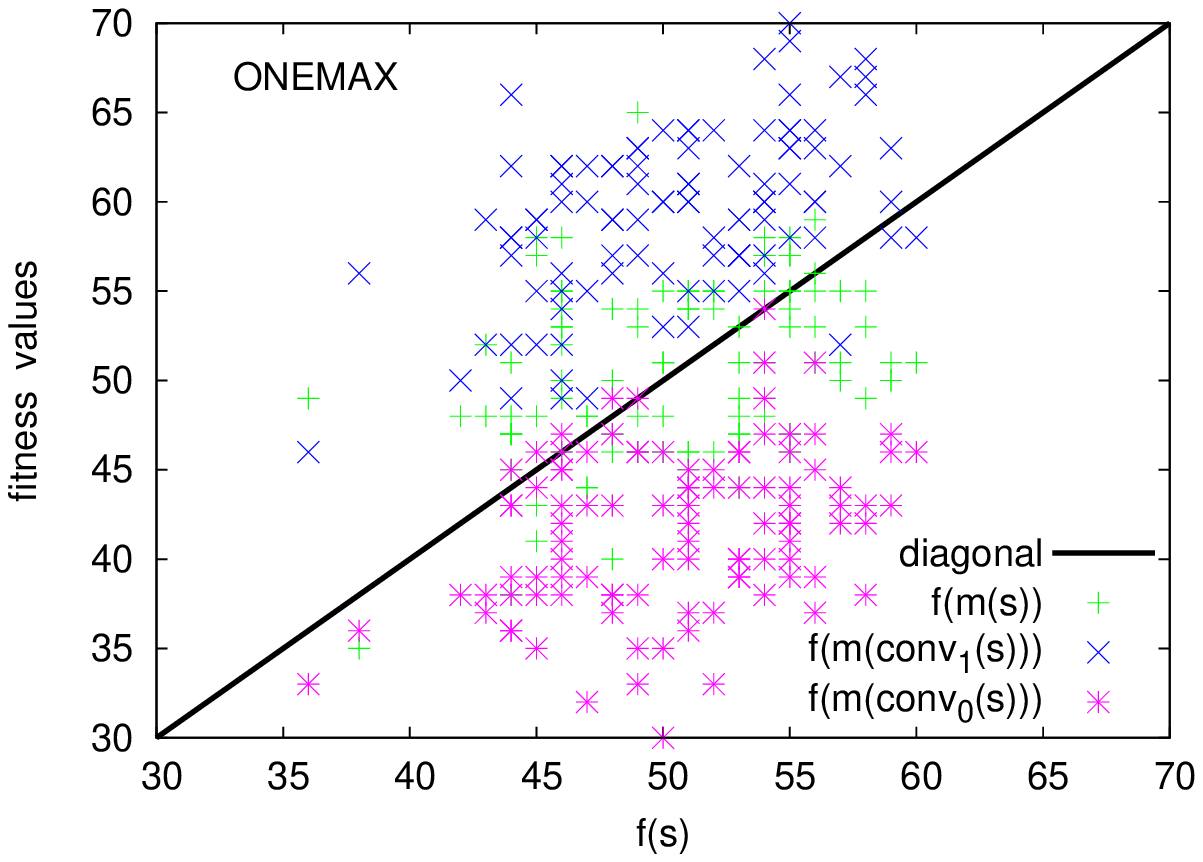} \\
\includegraphics[width=80mm,height=50mm]{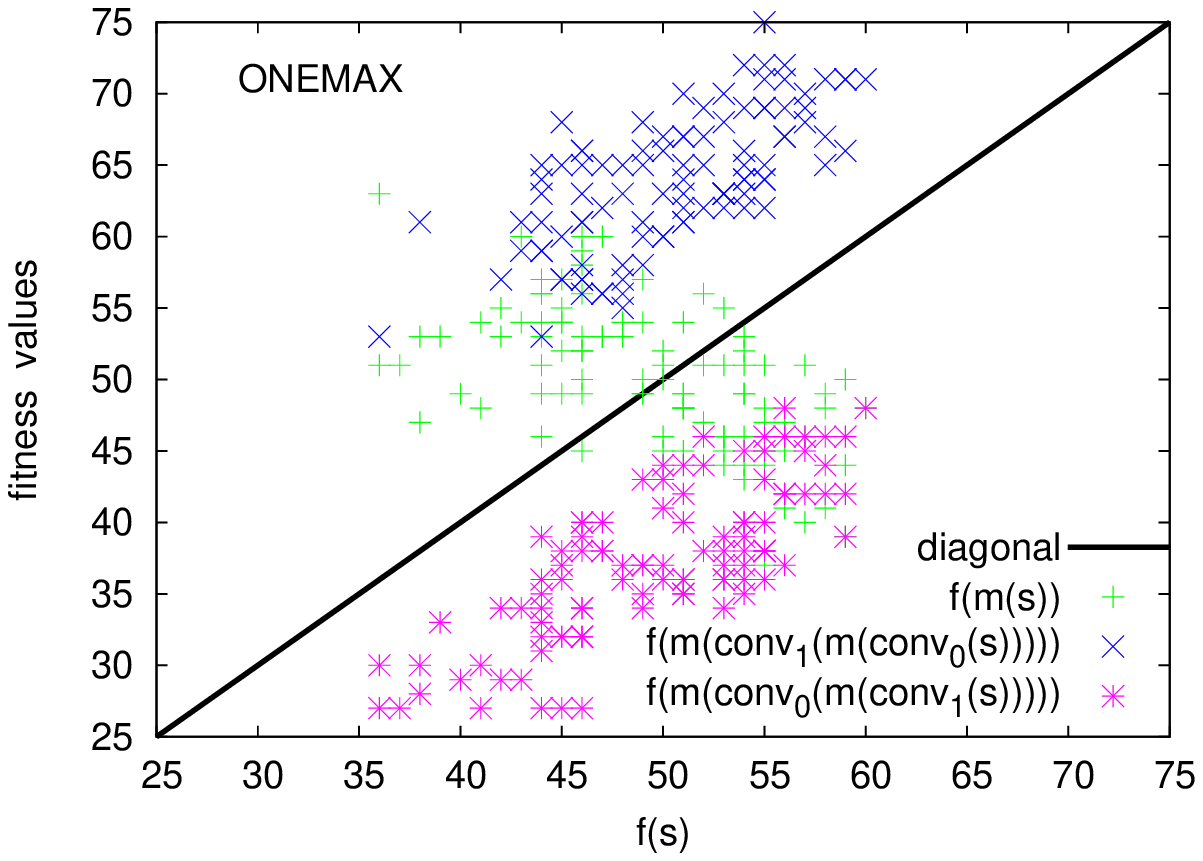} \\
\end{tabular}
\caption{Fitness Clouds representing the influence and interaction of BBC conversion routines with genetic operators: for the ONEMAX problem, we plot different kinds of fitness values for $100$ arbitrary solutions. The x-axis represents the fitness of initial solutions ($f$). The y-axis represents the fitness of solutions after the application of BBC conversion operators ($conv$) and the bit-flip mutation operator ($m$).}
\label{figGraphComp}
\end{center}
\end{figure}

\begin{figure}
\begin{center}
\begin{tabular}{@{}c@{}c@{}}
\includegraphics[width=42mm,height=30mm]{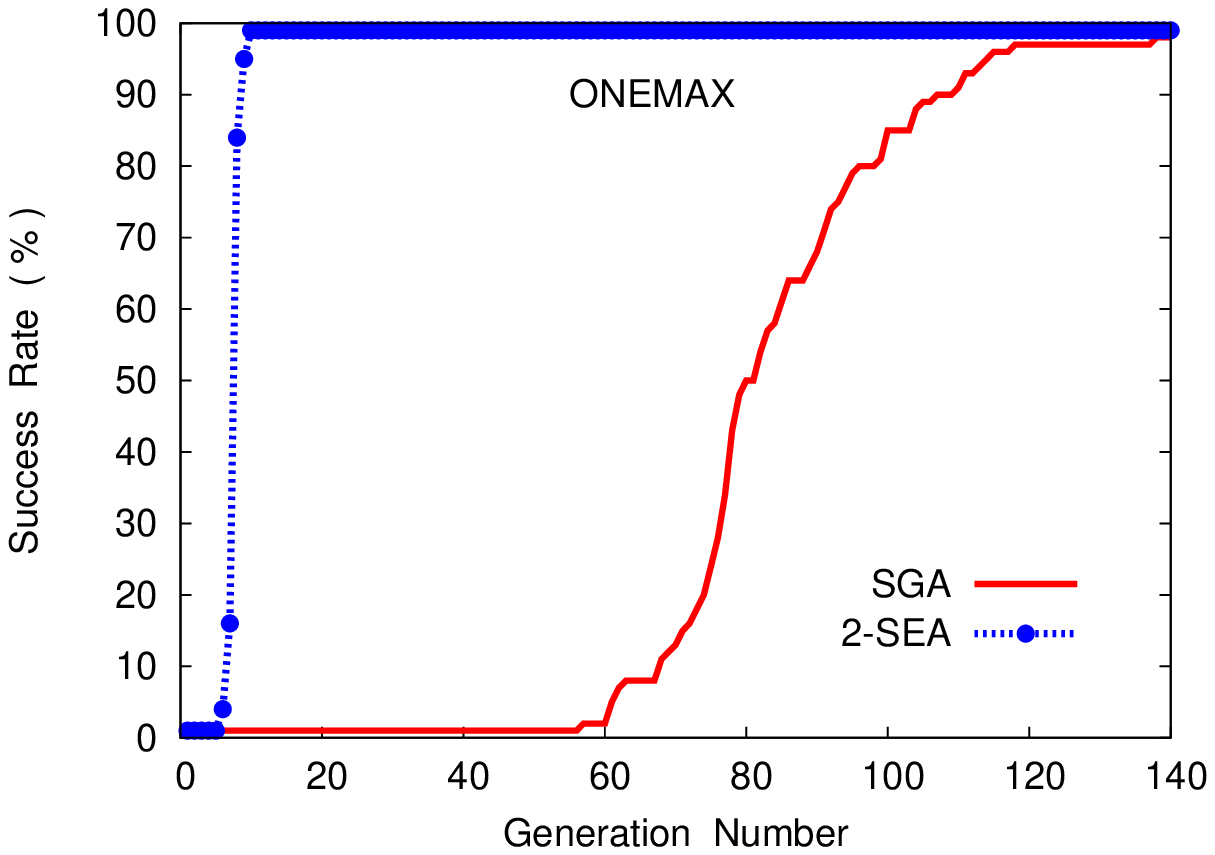} & \includegraphics[width=42mm,height=30mm]{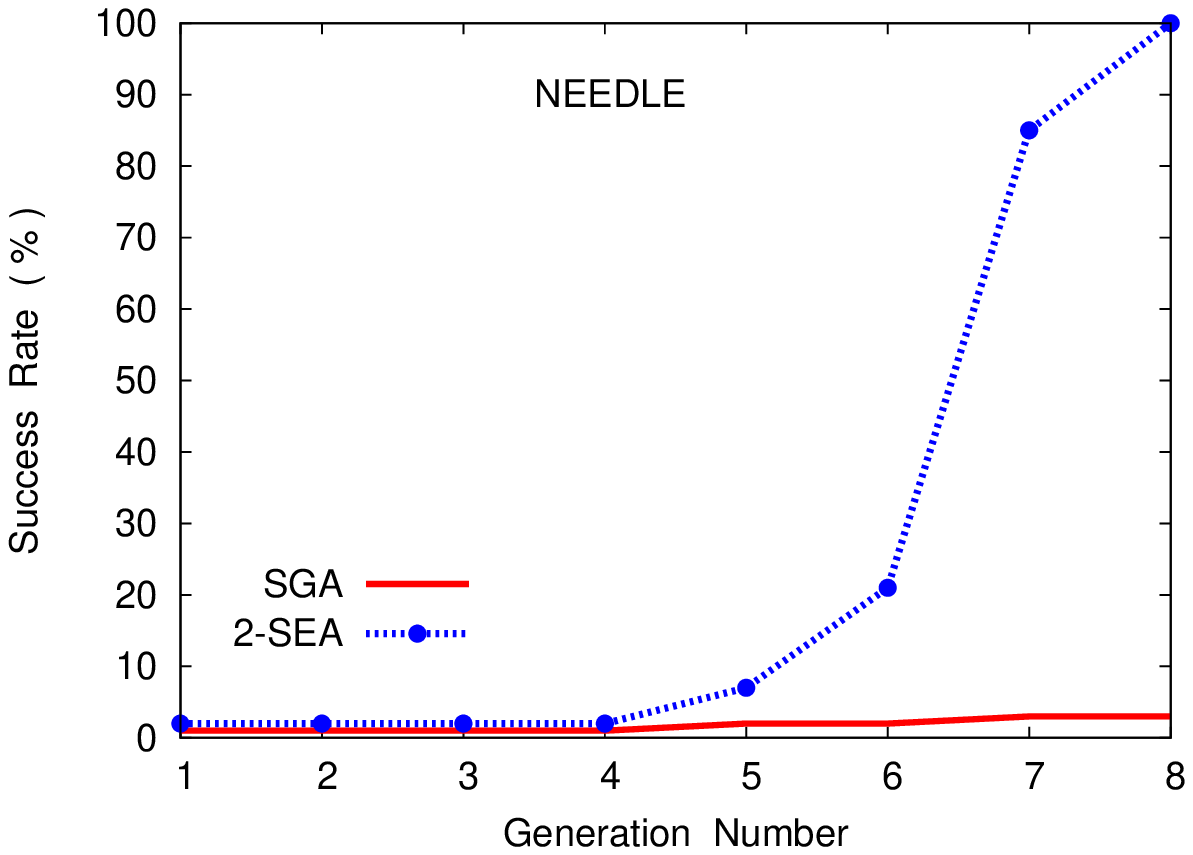} \\
\includegraphics[width=42mm,height=30mm]{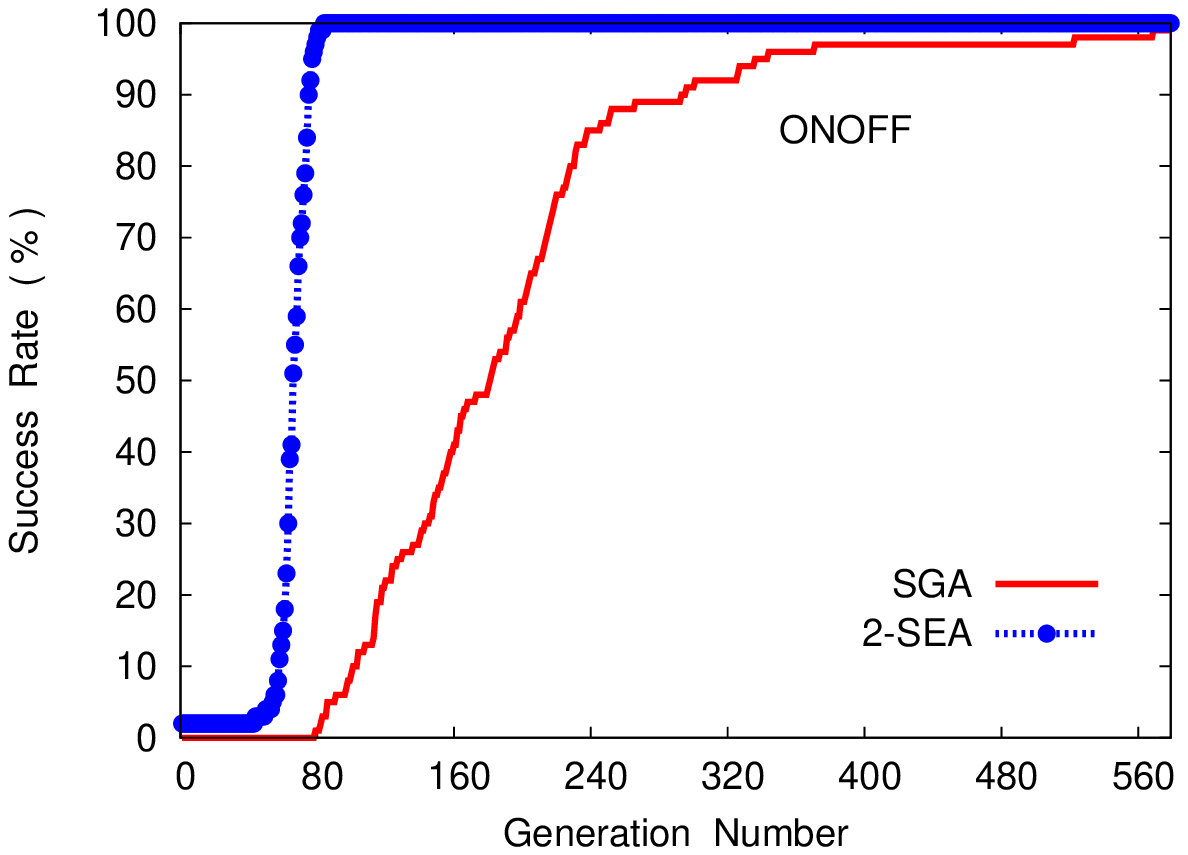} & \includegraphics[width=42mm,height=30mm]{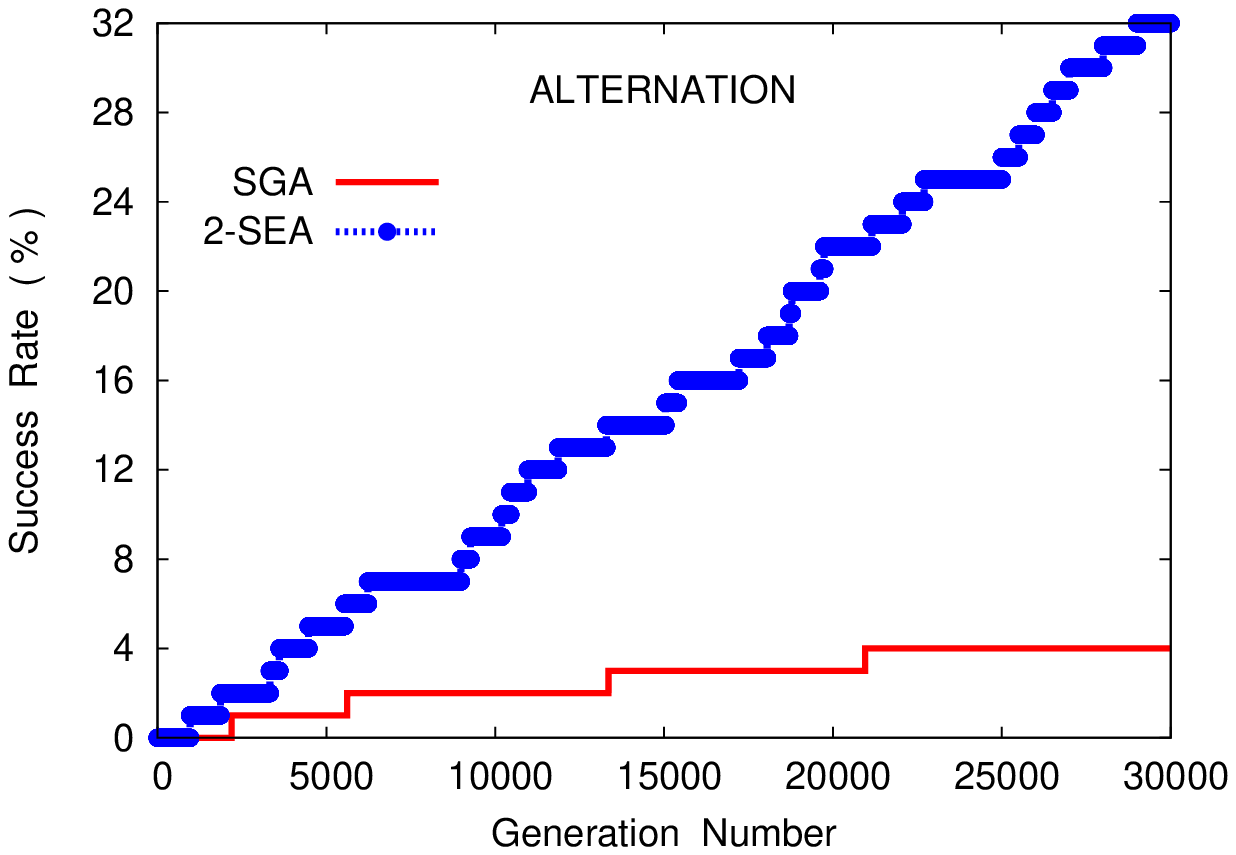} \\
\end{tabular}
\caption{Performance comparison of the percentage of correct solved solutions across the number of iterations required to reach the global optimum. These records were averaged over $100$ independent runs for each test function.}
\label{figExpRes}
\end{center}
\end{figure}

In this paper, we used a method based on the framework of change of the representation during the search. The basic intermediators to apply this action were the diverse BBC conversion operators. These operators allow to alter the representation of solutions from one coding to another during the search without modifying their respective fitness values. For this purpose, we applied identical representations of BBC which have the same search space, the same neighborhood structure and the same fitness values for identical solutions, but various conversion operators to change the form of the representation issued from BBC. We have to state that all previous works which used to search for the good coding during the optimization process and then tried to apply that best coding were very essential, helpful and efficient. Another effective statement and affirmation can be deduced from our work and test results. The data of the experiments shown in Table \ref{tabExpRes} and Figure \ref{figExpRes} clearly prove the importance and the utility of changing the representation of many random solutions in favour of 2-SEA that incorporates a conversion strategy which leads to a dynamic and mutual representation. They also confirm that the change of the representation during the search is a very helpful and fundamental step to profoundly think about as well to apply when one tries to improve EAs performances. \\
Besides, the experimental results displayed in Table \ref{tabExpRes} and Figure \ref{figExpRes} are uncomparable and show the advancement of 2-SEA over {\small S}GA. They distinctly show how 2-SEA has found the global optimum in an extreme minimal number of generations for the first three test functions while {\small S}GA has reached the global optimum of the ONEMAX and ONOFF problems in a remarkable larger number of iterations and failed to detect the global optimum of the NEEDLE and ALTERNATION problems for the majority of runs (for a great proportion of initial populations). For the deceptive P4 problem, the change of the representation with an appropriate conversion rate has driven the search process in 2-SEA to build and fix each bit in its correct position relatively to its neighbours but good combinations of bits cannot be made fast enough because of the matter that the bits are tightly linked each to other. Consequently, the change of the representation during the optimization task has been proved to be of great importance in EAs operation and positively showed that the obtained results for 2-SEA are significantly different from those of {\small S}GA for all objective functions. \\
To prove our results, first we must show value of applying the coding conversion operators. Thus, to reveal some characteristics of BBC and study the evolvability of $conv_0$ and $conv_1$, we have made a simple test on the ONEMAX problem denoted {\it Fitness Clouds Representation} \cite{r43}. We started our test with a fixed number of arbitrary solutions uniformly generated from a given $seed$ number. In a first step, we applied a standard bit-flip mutation to each of those solutions and evaluated their respective fitnesses ($m$). As a next step, we applied two kinds of coding conversion to those initial solutions, the first is done using $conv_1$ operator and the second using $conv_0$ operator. Then, we applied a standard bit-flip mutation to each of those solutions and evaluated their respective fitnesses ($m \circ conv_1$ and $m \circ conv_0$). In a following step, we applied two types of coding alternation ``tour'' to the same random solutions taken before, where each ``tour'' is considered as two consecutive coding conversions. The first ``tour'' is realized according to the respective application of $conv_1$, a bit-flip mutation, $conv_0$, a bit-flip mutation, and the evaluation of the corresponding fitnesses ($m \circ conv_1 \circ m \circ conv_0$). Inversely, the second ``tour'' is realized according to the respective application of $conv_0$, a bit-flip mutation, $conv_1$, a bit-flip mutation, and the evaluation of the corresponding fitnesses ($m \circ conv_0 \circ m \circ conv_1$). This elementary test was performed on $100$ arbitrary solutions, each having a length of $1900$, an extreme value of $k$ equal to $19$, a blocks number equal to $100$, and hence the ultimate solution is an all ``$1s$'' string with a fitness value equal to $100$. The traditional bit-flip mutation operator was applied in all steps with a bit-flip mutation rate equal to $0.25$. The comparison of the obtained results is illustrated in Figure \ref{figGraphComp}. \\
A graphical interpretation of Figure \ref{figGraphComp} (left) indicates that the fitness values of individuals which have been submitted to $conv_1$ and then to the bit-flip mutation operator are higher than those of individuals which have been simply submitted to a bit-flip mutation and than those of individuals which have been submitted to $conv_0$ and then to the bit-flip mutation operator. Similarly, Figure \ref{figGraphComp} (right) shows that the fitness values of individuals which have been submitted to $conv_0$ and then to $conv_1$ and then to the bit-flip mutation operator are higher than those of individuals which have been simply submitted to a bit-flip mutation and than those of individuals which have been submitted to $conv_1$ and then to $conv_0$ and then to the bit-flip mutation operator. Since for the ONEMAX problem, the more the number of ``$1s$'' in the string increases the more the corresponding fitness value increases, Figure \ref{figGraphComp} proves very well that $conv_1$ is the most appropriate conversion operator and is the one that clearly contributed in producing superior results. We can conclude that the coding alternation ``tour'' and the conversion of the representation from one coding to another have induced a befitting evolvability that matches to the problem structure. And the test results assume that the last applied BBC conversion operator is the one that influences the more on the final outcome. Graphical representations of fitness variations relatively to $pMutState$ given in Figure \ref{fig-er-functions} showed that, for all test functions, a large conversion rate is needed for 2-SEA to render high positive results, the fact that justifies once a time the important and essential role of BBC coding alternation ``tour'' and its constructive influence on the performance of 2-SEA by re-creating, remodeling and reforming the meaningful building blocks. Also, we can say that the evolvability of a coding conversion operator is more beneficial over the EAs performance after the change of the representation. The use of more than one coding in EAs is very important and the most fundamental mechanism resides in the framework of changing the representation from one coding to another which contributes in exploring undetected and unspoiled sub-regions of the search space. Therefore, better fitness values can be discovered and EAs can progress towards more positive outcomes. The test results have distinctly verified the utility of BBC and the value of coupling various encodings in an alternation strategy where different conversion operators interact to increase the probabilities of obtaining advanced and good structures. \\
As a final statement, we believe that our test results totally complied with our assertions about the argument ``Do not Choose Representation just Change'', and showed that an algorithm which incorporates a method of coding mating by the application of the conversion of the representation from one coding to another during the search will easily converge and will be more successful in reaching more optimum solutions using less computational power.

\section{Future Directions}
\label{secFD}
In our study, we used two {\small S}GAs in 2-SEA. An advanced research can lead to the exploitation of other kinds of EAs to be assigned to each state having in mind that both the notion of states and the state conversions are very essential in EAs functioning. \\
In this paper, BBC is considered. Though, other kinds of coding schemes such as tree or linear representation, and any number of coding schemes can be applied to EAs in order to profit from the convenient representation for a particular problem. \\
A future direction also suggests that other implementations of the SEA can still be improved by decreasing user defined parameters and making them automatically adjustable based on measures extracted from the process. \\
In a further research, we must understand properly the basic properties of BBC and recognize well its fundamental evolvability evoked by the genetic operators so we can propose other types of BBC conversion operators to help making the representation more dynamic and more adaptive to the problem structure.

\nocite{*}
\bibliographystyle{unsrt}
\bibliography{GECCO2009}

\end{document}